# AUTONOMOUS ROBOTIC BRIDGING USING DISTRIBUTED SWARM CONTROL WITHOUT INTER-AGENT COMMUNICATION

**Vishwaak C. Thamaraiselvan[1], Cody L. Lundberg[1], Michail Theofandis[1], Suhas Chelian[1], Nicholas R. Gans[1]**

[1]The University of Texas at Arlington Research Institute (UTARI), Fort Worth, TX

## ABSTRACT

*We describe SCARAB – Swarm-Capable Autonomous Robotic Aquatic Bridging. Using distributed swarm control and multi-model sensing of agents and docking targets, agents can localize themselves, join into formations and proceed to desired target locations. Our technologies would eventually allow the Army to perform unpredictable, dispersed river crossings, enhance crew survivability, and minimize the logistics footprint compared to the current Improved Ribbon Bridge. Our methods operate without GPS or RF communications, though these can be used in non-contested environments (e.g., civilian disaster relief for flooding, etc.). We demonstrate our system via physics-engine-based simulation of several agents using unmanned surface vehicles (USVs) in the presence of currents and wind.*

## 1. INTRODUCTION

The Army seeks autonomous robotic bridging for the 2040 battlespace (Army STTR A254-012; [1]). Current gap grossing technologies do not consider near-peer adversarial capabilities and support sustainment operations in a lethal contested logistics environment when the enemy can attack targets at virtually any depth within the battlespace. Autonomous, powered, floating bridges or rafts, however, will allow the Army to perform unpredictable, dispersed river crossings, enhance crew survivability, and minimize the logistics footprint compared to the current Improved Ribbon Bridge. University of Texas Arlington Research Institute (UTARI) in collaboration with Hydronalix, Inc. developed SCARAB - Swarm-Capable Autonomous Robotic Aquatic Bridging - to explore system components for the Army's need for autonomous robotic bridging. UTARI focused on two aspects: 1) distributed nonlinear formation control based on the work of [2] with new boat models from Hydronalix, and 2) multi-modal sensing and localization for unmanned surface vehicles (USVs) and docking targets. SCARAB has been demonstrated via simulation (Figure 1) and prior work has demonstrated our swarm technology on unmanned aerial vehicles (UAVs) and unmanned ground vehicles (UGVs) with different control strategies (unicycle, and front- or rear- wheel drive). Our technologies can function independently of GPS and RF communications, though these can be incorporated in environments which are not contested such as civilian disaster-relief scenarios for flooding, search and rescue, wildlife monitoring, etc.

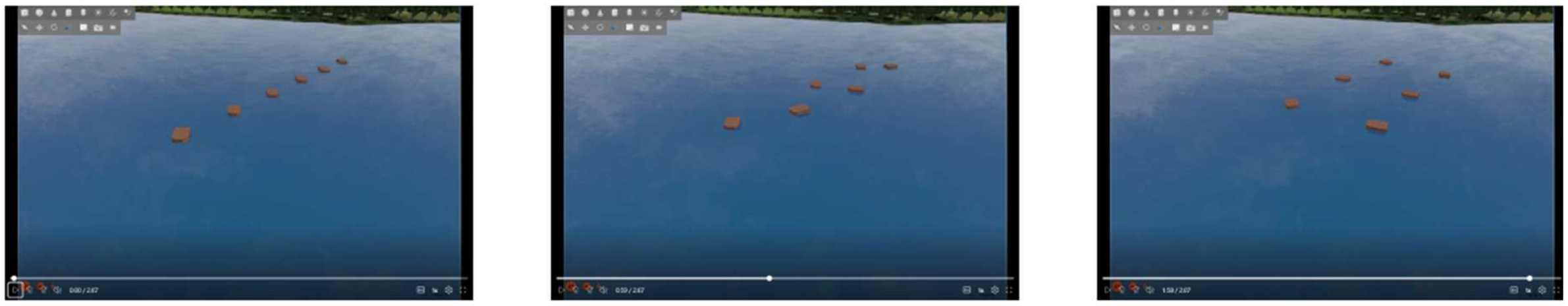

(a) Initial position at sim time= 0 (b) Snap shot at sim time= 227 (c) Final formation sim time= 482

**Figure 1:** Progression of the swarm simulation from initial deployment to final formation for a 2x3 "rafting" formation to carry large loads. Autonomous bays do not use GPS or inter-agent communication (e.g., RF) to reach their final position. This is especially useful in contested environments but not necessary in other applications (e.g., civilian disaster relief missions for flooding, etc.).

## 2. BACKGROUND

Many methods of formation control rely on global measurements such as GPS. Consensus-based methods, or techniques based on distributed pose, do not require global sensing. Nevertheless, vehicles must communicate with their peers during missions to estimate their pose, synchronize their orientation, or align their local coordinate frames relative to a common heading direction. As another alternative, distance-based or bearing-based formation control does not necessitate communication or common orientation, but they may converge to a local-minima that does not represent the correct formation.

Our approach utilizes barycentric coordinates, thereby eliminating these limitations. This is especially relevant to the Army because the ability to not use GPS and restrict RF communications in a contested environment which improves robustness and survivability. Optionally, our method can use these technologies where it is acceptable, e.g., emergency relief for flooding, etc. for civilian operations.

We provide a distributed, provably convergent, and robust formation control strategy for a wide range of vehicle dynamics such as USVs, UAVs, and UGVs. This approach negates the need for global position measurements, common heading direction, inter-agent communication, and a complete sensing graph required by existing formation control literature. Additionally, our method remains stable despite significant disturbances and input saturation. It incorporates a TRL 4+ fully distributed collision avoidance algorithm with stability guarantees that has been demonstrated on ground robots and drones.

Our work is based off the work of Fathian et al. [2], which presented a distributed control framework for achieving and maintaining planar formations in multi-agent systems with heterogeneous dynamics. The approach was built on barycentric-coordinate-based (BCB) control, in which each agent regulates its position relative to the position of its neighbors in the formation. This allowed fully distributed formation control without global position information or centralized coordination. By relying solely on local relative measurements, the method enforces a prescribed geometric formation while remaining scalable and robust to network size and topology. A key contribution of their work is the extension of BCB-based formation control to accommodate agents with different dynamic models and to improve robustness to modeling uncertainties and disturbances. Their work was demonstrated for wheeled ground vehicles, including car-like and tank-like vehicles, and aerial vehicles, including quadrotors and fixed-wing aircraft.

Fathian et al. [2] analyzed stability properties of the resulting closed-loop system and demonstrated that the desired planar formation can be achieved under broad conditions on agent dynamics and interaction structures. This robustness-focused formulation broadens the applicability of distributed formation control to realistic multi-agent settings where idealized dynamics and precise models cannot be assumed. It was also proved to be extremely robust to outside disturbances or need to change trajectory for collision avoidance. In simulation and with ground robots, they showed swarm formation of up to nine units.

Localization of units was shown using multiple cameras around the units but can also be adapted to use fiducials such as AprilTags, pose estimation by template-based methods (e.g., [3]), or scenic landmarks [4]. AprilTags are fiducial makers that provide six degree of freedom pose information ($x$, $y$, $z$, and rotation angles abouts these axes). This can be used to localize agents or docking targets. We have shown that AprilTags can be localized at more than 30 meters with less than 5% error, in day or night.

## 3. APPROACH

### *3.1. Simulation software and multi-modal sensing of agents and docking targets*

Our initial simulation platform is VRX (Virtual RobotX) simulator [5]. This simulator is used by the VRX competition and is built on top of Gazebo, which interfaces with Robot Operating System (ROS). Gazebo supports several features such as several different surface vehicles, modeling of hydrodynamic forces (including added mass, water currents, linear and quadratic damping), surface waves, thruster models, etc. The VRX simulator includes the WAM-V (Wave Adaptive Modular-Vehicle) USV pre-configured with multiple cameras, GPS, IMU, 3D LIDAR, and other sensors.

Hydronalix's Reckless USV frame was combined with the WAM-V model, the typical model for the VRX competition. This involved importing the Reckless 3D mesh for visuals and collisions and modifying the dynamics and propulsion. Steering control gains and max steering angles were also updated. Figure 2 shows multiple USVs spawned into the simulation. GPS-denied localization was developed via AprilTag computer vision. Figure 3 shows the USV with a camera pointed towards the AprilTag cube that acts as a docking target. RVIZ (Robot Visualizer) output is shown in Figure 4.

LIDAR was used as a baseline to ascertain the accuracy of camera-based localization of AprilTags. Figure 4 displays the centroids estimated for the USV, 3D LIDAR point cloud, and AprilTag detection via computer vision. The LIDAR data is shown as a cloud, and the AprilTag centroids are shown as small coordinate reference frames. This figure shows that the cube estimated surface and tag estimated centroids are co-planner and within 5% error.

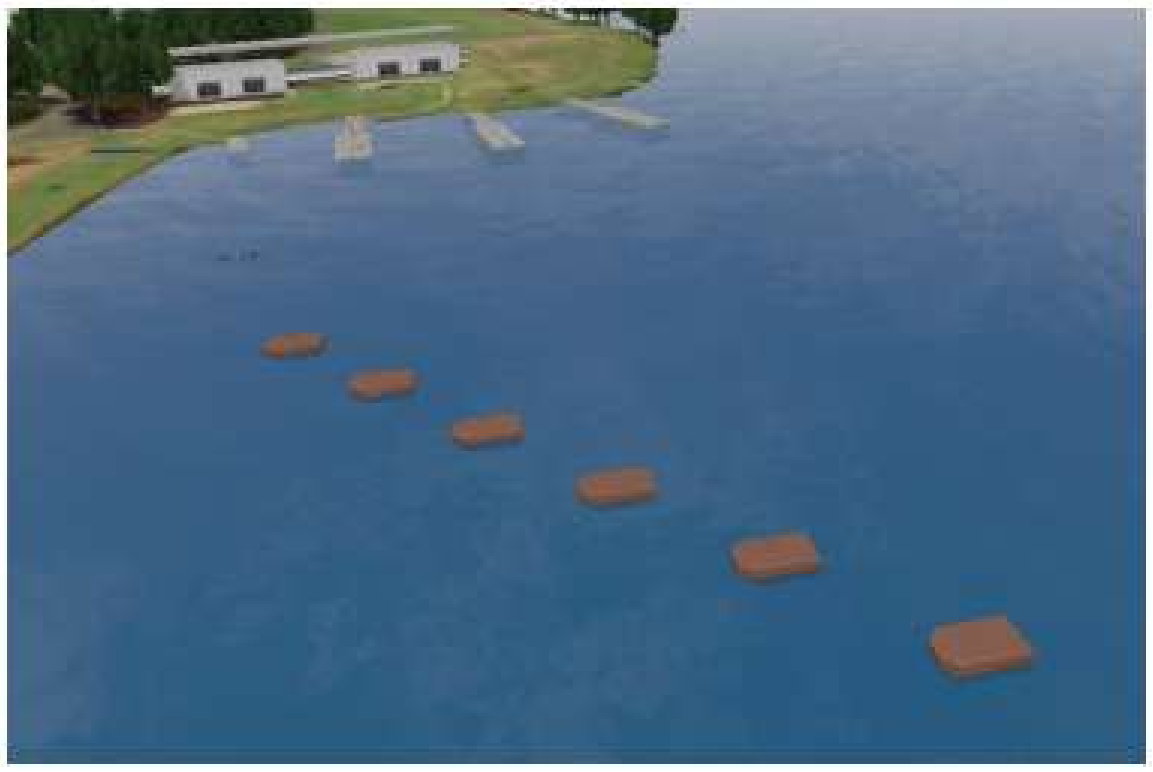

**Figure 2:** Multiple USVs with Hydronalix Reckless frames in the VRX simulator. UTARI imported Hydronalix's 3D model. Tuning of the propulsion system and dynamic model remains as future work.

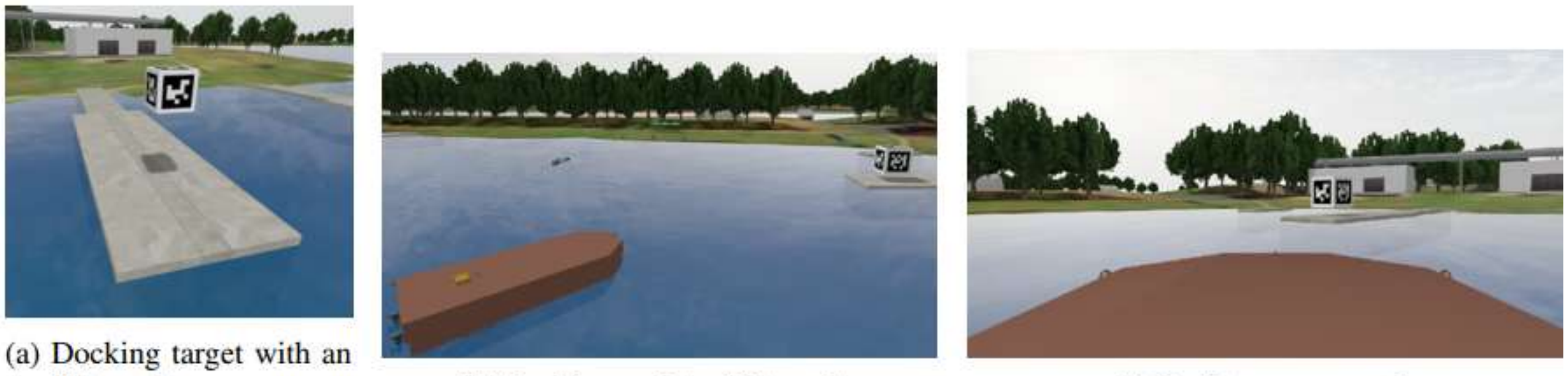

(a) Docking target with an AprilTag cube (b) Reckless and AprilTag cube (c) Reckless camera view

**Figure 3:** USV and AprilTag cube in the VRX simulator. Static AprilTag cubes are used in the world as an additional localization method in GPS degraded or denied areas. The AprilTag cube can be seen in the world view and the USV camera view. In future work, AprilTags can be replaced by advanced computer vision routines that can localize arbitrary docking targets (e.g., "find the tree, go 5 degrees to the right of it", etc.)

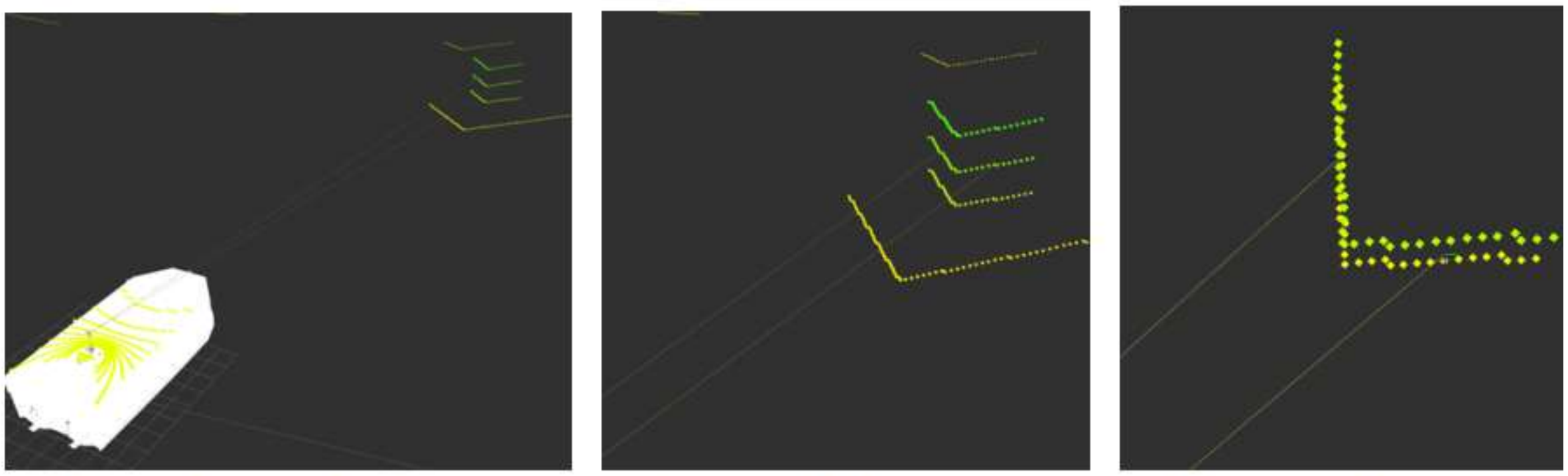

(a) Reckless boat and LIDAR point cloud (b) Point cloud of AprilTag cube (c) Point cloud and tag centroids

**Figure 4:** Reckless USV, 3D LIDAR point cloud, and detected AprilTag centroids are all shown using RViz. The AprilTag centroids are detected via the onboard camera, and their positions are compared against the LIDAR. point cloud. The image above shows that the AprilTag centroids lie on the surface of the cube as seen by the LIDAR and both diverge by less than 5%. These separate sensor streams are showing agreement in the cube location.

Figure 5 shows the visualization of the fused localization (described below). There are two defined frames for the AprilTag cube, one is statically defined in the world coordinate frame and one is estimated from the AprilTags. When multiple AprilTags are in view, the center of the cube is estimated by applying known offsets and combining the multiple tag estimations with spherical linear interpolation. Exponential weights were chosen to balance quick state updates and while minimizing outlier positions disturbances. There is a defined 90 degree offset between the two locations shown for the cube, they are nearly concentric. Three positions are displayed for the USV: one from the GPS signal, one is calculated by reversing the AprilTag estimation and applying it to the know static AprilTag cube location, and finally the fused positions is calculated from creating a short buffer holding both GPS and reversed AprilTag estimates then an exponentially weighted moving average is applied to this buffer to provide a combined location of the USV.

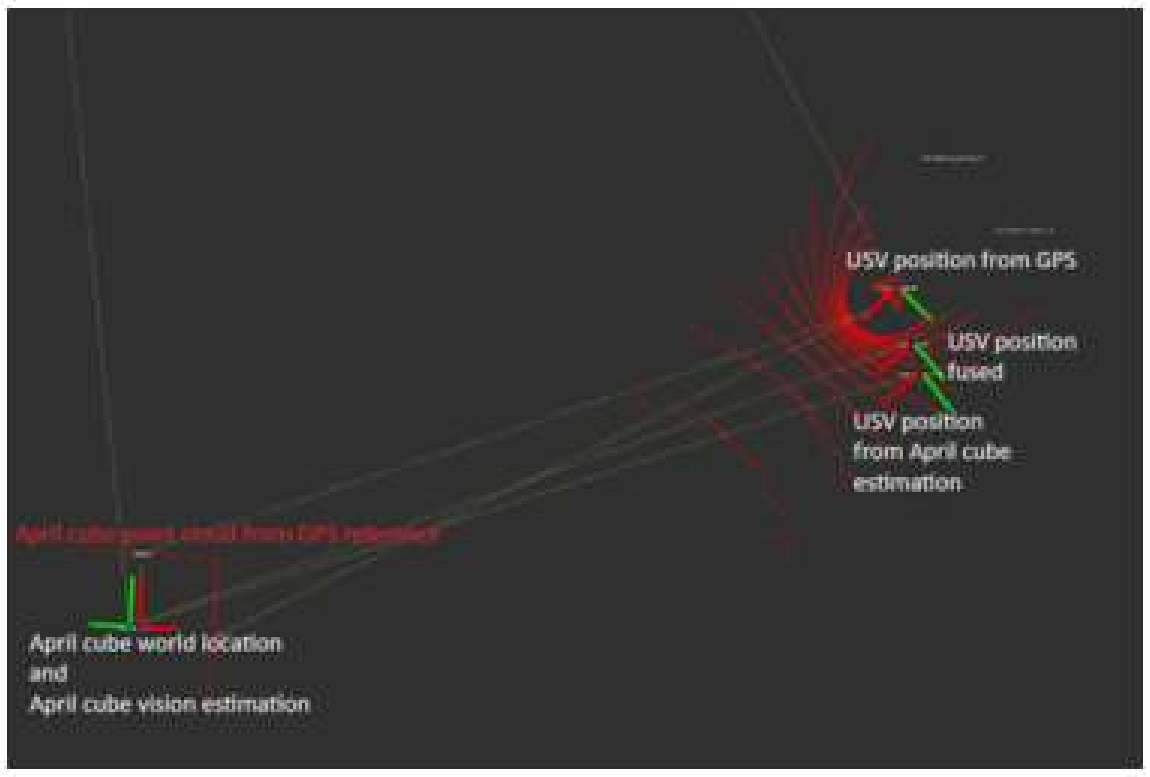

**Figure 5:** This figure shows the fusion of the USV GPS and AprilTag cube localization. Both the known static position of the AprilTag cube from the world frame and the estimated cube position are show at the bottom left. The outline of the AprilTag cube is overlaid from the 3D LIDAR data. A new localized USV position is calculated by fusion the GPS and AprilTag localization methods.

### *3.2. Vehicle control model*

We have extended the approach in [2] to incorporate the dynamics and control of surface vessels, as well as common disturbances such as wind or river currents. Figure 6 shows our control architecture for the USV operating under a formation control framework. The formation controller generates reference commands for both heading ($c_\delta$) and speed ($c_v$), which are sent to separate heading and speed controllers. These controllers compute the control inputs—steering command ($\delta$) and velocity ($v$)—that drive the USV. The vehicle is actuated by two podded thrusters and is equipped with GPS and IMU sensors, which provide feedback on its position, velocity, and orientation. GPS is not necessary but can be used when it is available. This feedback is used to compute tracking errors in heading ($e_\delta$) and speed ($e_v$), which are fed back into the respective controllers to ensure accurate tracking of the desired formation behavior. The dashed lines represent the feedback loops that enable the USVs to maintain coordination and stability within a formation despite disturbances. Figure 7 shows simulation data of how the controller rejects wave and wind disturbances to closely follow a straight-line path.

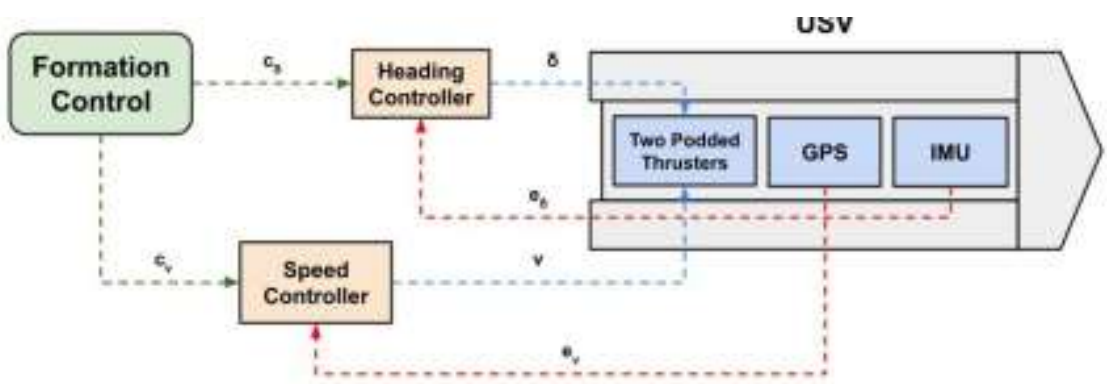

**Figure 6:** USV feedback control system block diagram. Each simulated USV in Gazebo is equipped with two podded thrusters, an IMU, and a GPS sensor. Our current control architecture employs two separate controllers: one regulates the vessel's heading using orientation data from the IMU, while the other controls its surface speed based on position and velocity information derived from GPS.

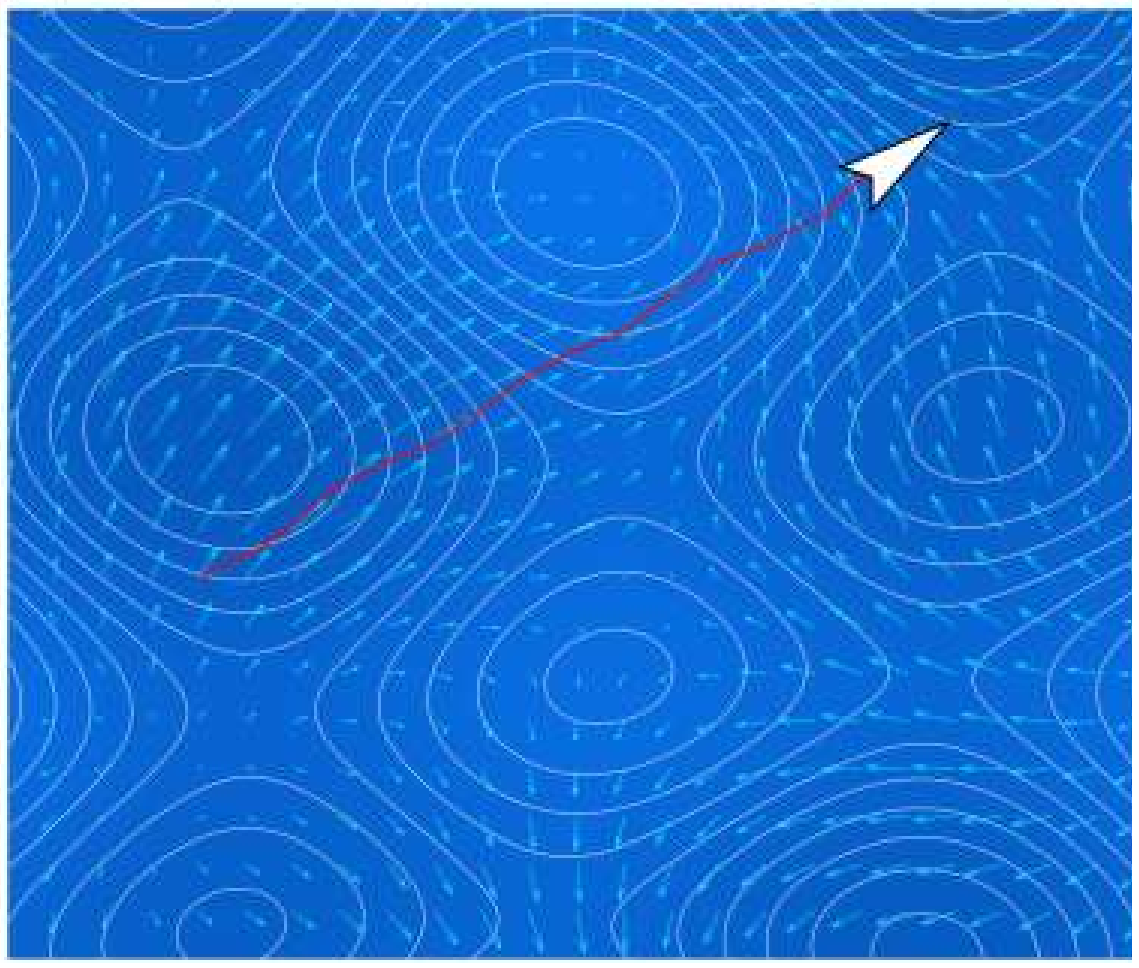

**Figure 7:** The figure illustrates the trajectory (in red) of the USV navigating the surface of a body of water subject to wave and wind disturbances. The blue background represents the sea surface, where the white color contours indicate the repulsion of the waves. Wind disturbances are visualized using cyan arrows, indicating both the direction and magnitude of the wind field across the lake. Velocity information derived from GPS.

### *3.3. Swarm formation control*

Coordinating USV swarms presents unique challenges due to environmental disturbances such as waves and wind, compounded by the absence of a shared global reference frame among agents. The underlying control strategy [4] requires neither inter-agent communication nor knowledge of absolute position, nor a shared global orientation. Agents operate using relative positions obtained in the local coordinate frame from an onboard vision sensor. This eliminates the need for GPS or RF-based communication. The controller further incorporates an active collision-avoidance system to ensure inter-agent separation. Its robustness to dynamic uncertainty makes the controller adaptable to various systems with different dynamics, for example, surface vehicles with hydrodynamic characteristics that differ from those of the ground and aerial vehicles on which the framework was originally validated. We extend this framework to a swarm of USVs and validate formation convergence in a simulation environment

We use two metrics to quantify the convergence of the swarm to the desired formation: the convergence time $e_t$ and mean formation error $e_f$. The mean formation error is defined as the average deviation of the inter-agent relative positions from their desired values across all agent pairs using Euclidean distance:

$$e_f = \frac{1}{|\mathcal{E}|} \sum_{(i,j)\in\mathcal{E}} \left\|(p_j - p_i) - (p_j^* - p_i^*)\right\| \quad (1)$$

where $p_i$ is the position of the agent $i$, $p^*_i$ is its desired position and $\varepsilon$ denotes the set of agent pairs for the formation. The individual formation error for agent $i$ is given by:

$$e_{f,i} = \left\|(p_j - p_i) - (p_j^* - p_i^*)\right\| \quad (2)$$

for each $(i, j) \in \varepsilon$. The convergence time $e_t$ is defined as the first-time instance $e_f < \epsilon$ remains below a desired tolerance for 10 consecutive simulation time steps for all $i$. We use $\epsilon = 1\ m$ for proof of concept and can reduce this tolerance in future work

## 4. EXPERIMENTS

All experiments were conducted in the Gazebo Simulation environment using ROS 2 Jazzy on an Intel i7 machine. A swarm of six USVs was simulated using the Hydronalix Reckless platform. We evaluate the formation convergence for two target configurations: a 2x3 "raft" formation and a formation in the shape of an 'X'. These are shown in Figure 1 with inter-agent spacing of 1 $m$. As no prior work has applied this framework to USVs, this constitutes the first validation of the controller in a maritime setting.

Figure 8 displays the formation error $e_{f,i}$ of each agent over time as well as the average over all agents, $e_f$. By step 482, all agents satisfy the convergence condition. This gives us $e_t$ = 492 time steps, which corresponds to 49.2 seconds in simulation time. Each simulation time step corresponds to $\Delta t = 0.1$ seconds of simulation time, with a real time factor of $k = 0.15$, which comes to 5 mins 49 second of wall clock time. Similarly for the X configuration as shown in Figure 9, the agent converged in $e_t$ = 332 with a wall time of 3 mins 59 seconds and simulation time of 33 seconds. These results highlight that we successfully used our vehicle model, swarm formation algorithms, and our simulation software to demonstrate autonomous raft formation of USVs.

Figures 9 and 10 show an alternate swarm control result. Here five bays start in random positions and converge to an "X" pattern.

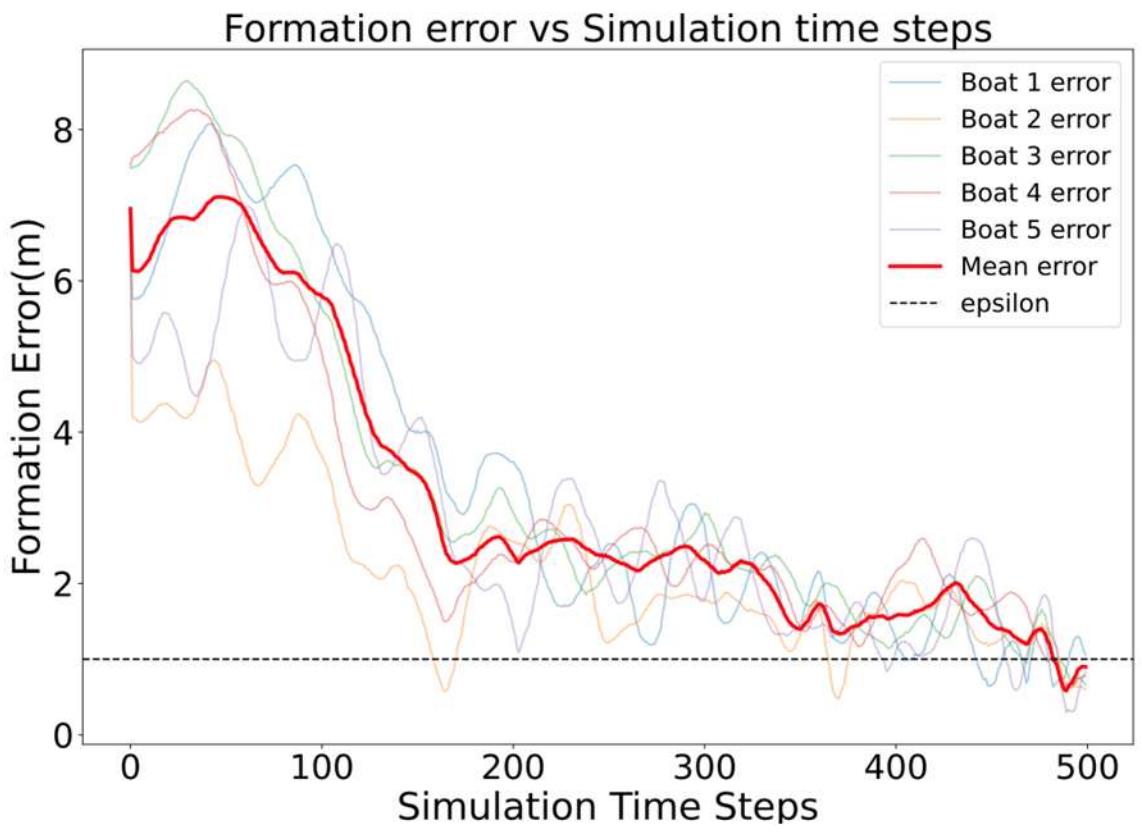

**Figure 8:** The red line indicates the mean formation error $e_f$ and colored lines indicate the per-agent position error vs simulation time step for the 2x3 formation. The dashed line indicates the convergence threshold.

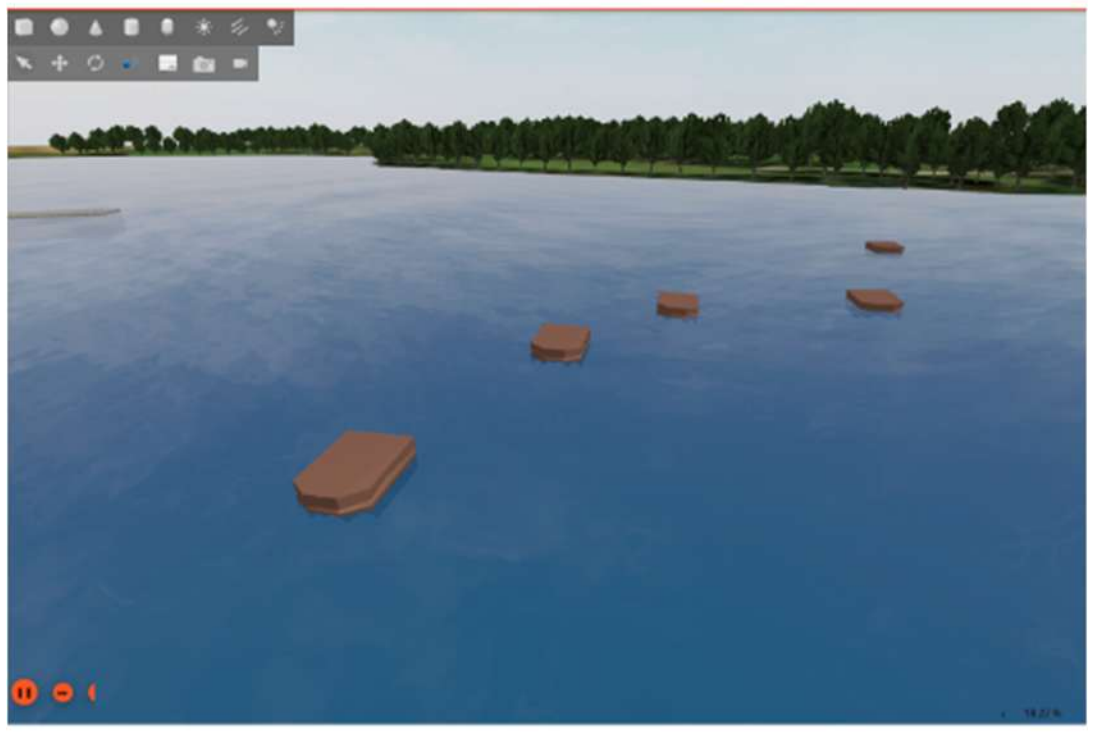

(a) Initial position at sim time = 0

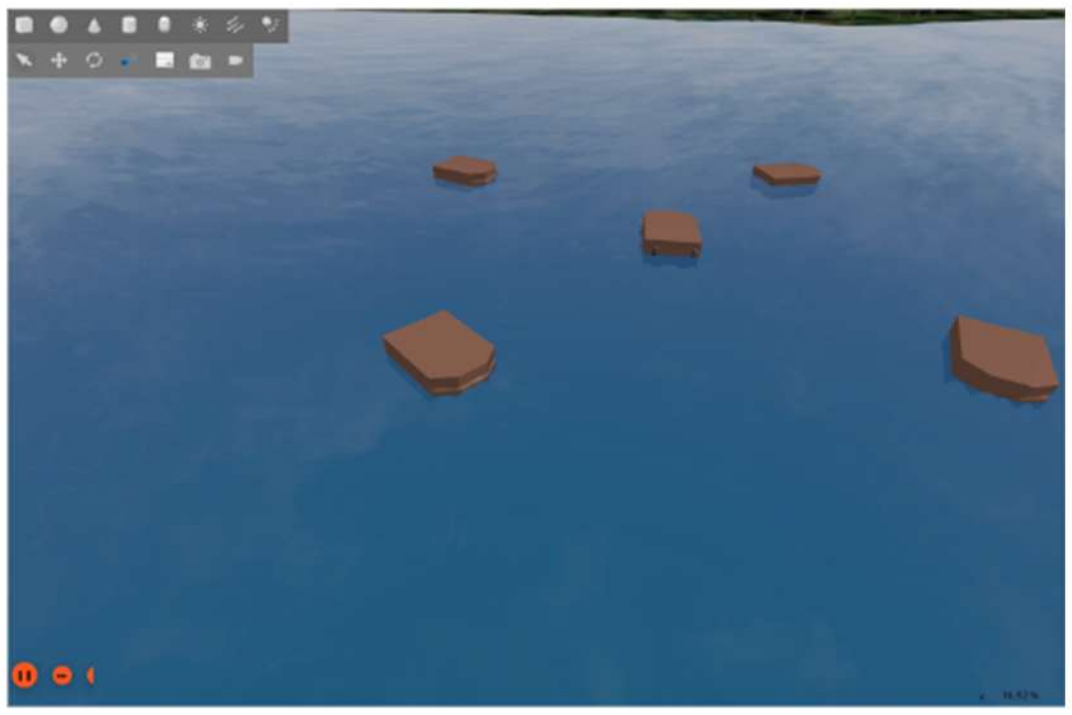

(b) Final position at sim time = 332

**Figure 9:** Progression of the swarm simulation from initial deployment to final formation for an X-shaped formation.

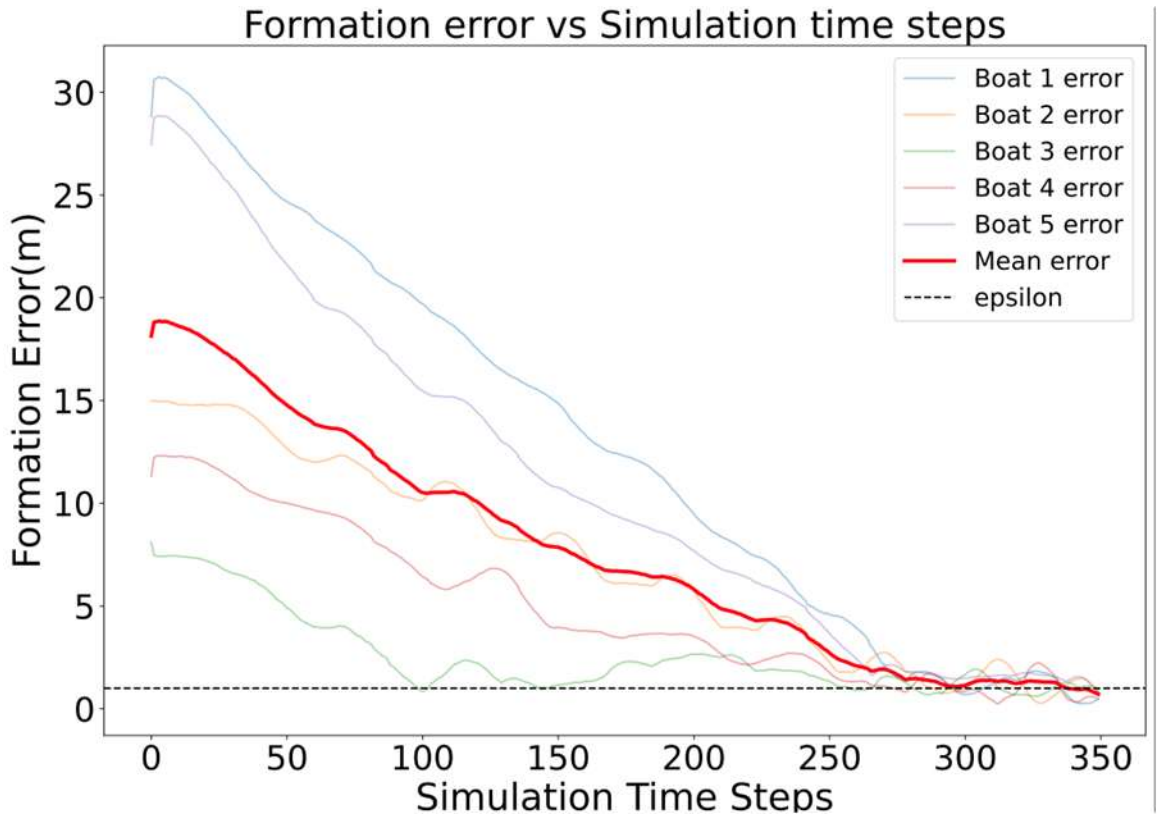

**Figure 10:** The red line indicates the mean formation error $e_f$ and colored lines indicate the per-agent position error vs simulation time step for X formation. The dashed line indicates the convergence threshold.

## 5. CONCLUSION

In this paper, we presented SCARAB - Swarm-capable Autonomous Robotic Aquatic Bridging for autonomous robotic bridging. Floating bays can assemble them autonomously using distributed swarm control, and multi-modal perception with optional communication via RF or acoustic modems (the latter is harder to jam).

There are several possible extensions to this work. For example, with regards to robustness, one could pursue: 1) overcoming obstacles such as terrain avoidance (e.g., sandbars, mines, etc.), 2) making sure vision perception modules are robust to adverse conditions such as smoke, fog or low light conditions, etc., and 3) formation control with damaged, disabled or non-responsive units. For 1, we have shown the ability to overcome obstacles using a potential fields approach [6]. For 2, we have shown the ability to detect arbitrary targets in day or night using an infrared camera and an embedded computer for low SWaP-C [7]. That work can also be completed by automatic detection of moving targets with background modeling [8]. For 3, one can detect disabled units if they do not move (or are not seen) after waiting a time period (e.g., 5 minutes). If they are non-responsive, additional bays can be used to create the desired formation (e.g., bring nine bays to the field when only six are needed; if bays 5 and 6 are not working, then use bays 8 and 9).

## 7. CONTACT INFORMATION

Nicholas R. Gans, PhD.
Professor of Computer Science and Engineering; Principal Scientist
The University of Texas at Arlington (UTA); UTA Research Institute (UTARI)
7300 Jack Newell Blvd. S
Fort Worth, TX
nick.gans@uta.edu
+1 817 272 5900

## 8. ACKNOWLEDGMENTS

Research was sponsored by Army STTR A25D-003 Autonomous Robotic Bridging (contract W51701-25-C-A253). We thank Hydronalix, Inc., the prime contractor, for their collaboration.